\documentclass{article}
\usepackage[final]{neurips_2023}
\usepackage[utf8]{inputenc} 
\usepackage[T1]{fontenc}    
\usepackage{amsmath}
\usepackage{hyperref}       
\usepackage{url}            
\usepackage{booktabs}       
\usepackage{amsfonts}       
\usepackage{nicefrac}       
\usepackage{microtype}      
\usepackage{xcolor}         
\usepackage{algorithm}
\usepackage{algpseudocode}
\usepackage{natbib}
\usepackage{graphicx}
\usepackage{placeins}
\usepackage{appendix}
\usepackage{subcaption}
\usepackage{longtable}
\title{Batch-Instructed Gradient for Prompt Evolution: Systematic Prompt Optimization for Enhanced Text-to-Image Synthesis}
\author{%
  Xinrui Yang\\
  University College London\\
  xinrui.yang.20@ucl.ac.uk\And
  Zhuohan Wang\\
  University College London\\
  zhuohan.wang.23@ucl.ac.uk\And
  Anthony Hu\\
  University College London\\
  anthony.hu.19@ucl.ac.uk\And
}

\begin{document}
\maketitle

\begin{abstract}
Text-to-image models have shown remarkable progress in generating high-quality images from user-provided prompts. Despite this, the quality of these images varies due to the models’ sensitivity to human language nuances. With advancements in large language models (LLMs), there are new opportunities to enhance prompt design for image generation tasks. Existing research primarily focuses on optimizing prompts for direct interaction, while less attention is given to scenarios involving intermediary agents, like the Stable Diffusion model. This study proposes a Multi-Agent framework to optimize input prompts for text-to-image generation models. Central to this framework is a prompt generation mechanism that refines initial queries using dynamic instructions, which evolve through iterative performance feedback. High-quality prompts are then fed into a state-of-the-art text-to-image model. A professional prompts database serves as a benchmark to guide the instruction modifier towards generating high-caliber prompts. A scoring system evaluates the generated images, and an LLM generates new instructions based on calculated gradients. This iterative process is managed by the Upper Confidence Bound (UCB) algorithm and assessed using the Human Preference Score v2 (HPSv2). Preliminary ablation studies highlight the effectiveness of various system components and suggest areas for future improvements.
\end{abstract}

\section{Introduction}
In recent years, text-to-image models \citep{ramesh2021zero, ramesh2022hierarchical, pryzant2023automatic, nichol2021glide} have demonstrated the ability to generate high-quality images from user-provided prompts. However, the quality of these images can vary significantly due to the models' sensitivity to the nuances of human language. With the rapid advancement of text generation models, particularly large language models (LLMs), we now have the tools to enhance prompt design for these image generation tasks. While existing research has extensively validated the capabilities of LLMs in optimizing prompts for direct interaction\citet{yang2024large}, less attention has been given to scenarios involving an intermediary agent, such as the Stable Diffusion model, where the prompt optimisation does not directly instruct the LLM but rather aims to improve the output of another system.

Previous studies have largely focused on optimizing "weight-accessible" language models using gradient descent  \citet{pryzant2023automatic}, an approach that, while effective, is both memory-intensive and time-consuming. Conversely, the most advanced LLMs are often "weight-inaccessible," accessible only via APIs, which presents unique challenges and opportunities for prompt optimisation.

In this study, we propose a Multi-Agent framework to enhance the capabilities of text-to-image generation models through the optimization of input prompts. Central to our framework is a prompt generation mechanism that intelligently crafts prompts from a repository of initial simple queries. These queries are selectively refined using a set of dynamic instructions, which are systematically evolved based on iterative performance feedback, refined iteratively by dedicated agents. Our iterative loop commences with a single initial instruction, which is employed by an instruction modifier, tasked with transforming initial prompts into sophisticated queries likely to yield higher-quality images. The resulting refined prompts are fed into a state-of-the-art text-to-image model to produce visual outputs.

To ensure the generation of high-fidelity images, we integrate a professional prompts database, an extensive collection of high-caliber prompts. This database functions as a benchmark, guiding the instruction modifier towards the quality of prompts that are proven to produce professional-grade images.

Subsequent to the image generation, a scoring system evaluates the outputs, separating the high-quality images from the subpar ones. Inspired by the Automatic Prompt Optimization (APO) method \citet{pryzant2023automatic}, the scores derived from this system are analyzed by a gradient calculator to ascertain the factors contributing to the prompt's success or failure. In a novel twist, we deploy another LLM to generate a variety of new instructions based on these calculated gradients. This introduces a breadth of potential modifications, ensuring that our instruction pool remains rich and versatile.

The feedback obtained from this scoring and generation cycle is used to update the instruction tracks, thus refining the instruction pool in a feedback loop. Additionally, to manage the instruction list size and to balance exploitation with exploration, we employ the Upper Confidence Bound (UCB) algorithm\citep{kuleshov2014algorithms}.

The entire process is informed by a sophisticated scoring function to accurately assess the quality of the generated images. We leverage the Human Preference Score v2 (HPSv2) \citep{wu2023human} as our scorer, which is claimed to be state-of-the-art in the field, trained on a substantial dataset of human preference data.
Through this rigorous and methodical approach to prompt optimization, our framework seeks to foster a more nuanced interaction between the linguistic understanding of LLMs and the visual creation capacity of image synthesis models. The objective is to evolve the system to a point where it can autonomously refine prompts to consistently produce images that rival professional standards, effectively narrowing the gap between human linguistic input and AI-generated visual content.

\begin{figure}[htbp]
  \centering
  \includegraphics[width=0.75\textwidth]{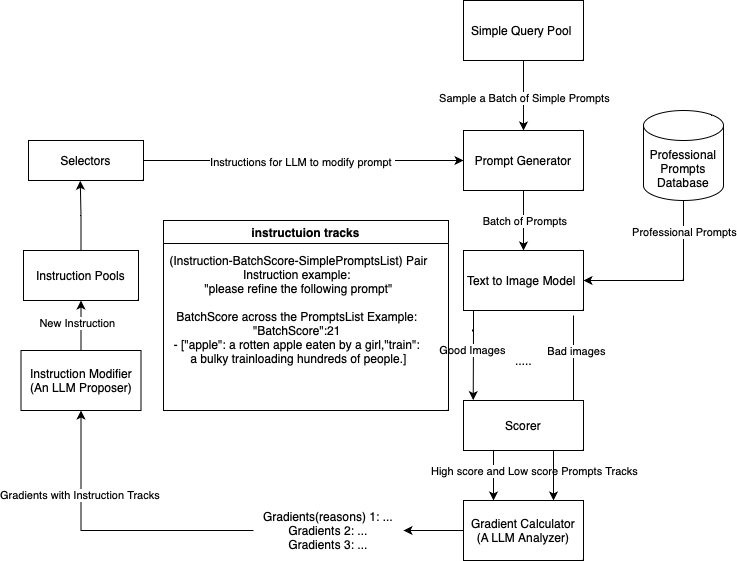}
  \caption{An Overview of the our Optimizing System}
\end{figure}

Due to time constraints within the scope of this study, we conducted a preliminary ablation study focusing on several pivotal components of our system. We scrutinized the selection method, the size of the query batch, and the influence of integrating a professional prompts database to assess their individual and collective impact on the image generation process. Our findings provide initial insights into the effectiveness of these components and highlight their respective contributions to the system's performance. In parallel, we identified potential weaknesses inherent in the current configuration, offering opportunities for future enhancements. These insights pave the way for subsequent, more in-depth research to further refine the system and maximize the quality of the AI-generated images, ensuring that our prompt optimization framework remains at the forefront of innovation in the field of text-to-image synthesis.

\section{Literature Review}
\subsection{Introduction to Current Text-to-Image Models and Prompt Optimisation}
Recent advancements in text-to-image models like Stable Diffusion XL \citep{podell2023sdxl} and Generative Adversarial Networks (GANs) have revolutionised the capability of generating high-quality images from textual prompts. These models often rely on nuanced prompt design to interpret human language effectively and produce relevant images. Large language models (LLMs) such as GPT and Llama are now extensively used to refine these prompts, thereby enhancing the quality of outputs. This literature review is going to discuss and compare different text-to-image generation approaches and how researchers use language models to optimize prompts and in turn improve image generation quality.
\subsection{Techniques of Prompt optimisation and Their Comparative Analysis}
\subsubsection{Reinforcement Learning and Supervised Fine-Tuning:}
The paper by \citet{hao2024optimizing} explores a hybrid approach. The approach is defined that initial prompt templates are refined using supervised learning to create a baseline which captures the intended stylistic and content-specific requirements of the images. They then utilised reinforcement learning algorithms adjust these prompts based on user feedback and model performance. The algorithm iteratively optimizes prompts to enhance both the aesthetic quality and relevance of the generated images. This method is particularly effective in scenarios where direct manipulation of model weights is not possible, offering a flexible and responsive approach to prompt optimisation. Our approach in this report has adapted this method since the weights of GPT are not accessible.
\subsubsection{Optimisation by Prompting (OPRO)}
\citet{yang2024large} study provides a novel use of LLMs as dynamic optimizers. LLMs refine prompts repeatedly, taking into account feedback from previous iterations This strategy, known as Optimisation by Prompting (OPRO), improves task accuracy dramatically by leveraging LLMs' natural language interpretation and generating skills. The iterative method is assessing the efficacy of a particular prompt, modifying it based on performance indicators, and returning the modified prompt into the system. This feedback loop enables continuous enhancement and adaptability. This increases the system's responsiveness to the intricacies of user input as well as the specialised requirements of text-to-image operations. The research demonstrates that this approach not only enhances image quality but also aligns more closely with user intentions. It in turn optimizes the interaction between human inputs and automated image generation systems.

\subsubsection{LLM-based Prompt Optimisations}
\citet{ma2024large} critically evaluate the operational issues associated with utilising LLMs to optimize prompts for text-to-image generation in their investigation of prompt optimisation with LLMs. They notice a key difficulty with LLMs' initial prompt outputs. The problem is these outputs are linguistically consistent but they are frequently ineffective in practical applications due to the stochastic nature of the model answers. To resolve these inconsistencies, the researchers propose a unique optimisation approach called "Automatic Behaviour Optimisation" (ABO). This method turns the focus away from typical prompt refinement and towards influencing the model's interpretative and generative processes. ABO aims to give a more steady and controlled method of optimising prompts, which is especially useful in applications that need accuracy and reliability. The application of ABO has demonstrated encouraging outcomes in terms of lowering output unpredictability and raising the produced prompts' usefulness, greatly enhancing the relationship between LLMs and subsequent text-to-image tasks.
\subsubsection{State-of-the-art Scoring Model}

The Human Preference Score version 2 (HPSv2) \citet{wu2023human} is a cutting-edge benchmarking tool for determining human preferences for images created via text-to-image synthesis. HPSv2 is based on the CLIP model and fine-tuned using a big human preference dataset. It outperforms earlier approaches including HPSv1 \citep{wu2023human1}, ImageReward \citep{xu2024imagereward}, and PickScore \citep{kirstain2024pick} for predicting which pictures people would like. It stands out for its excellent accuracy, which reached 83.3\% on the HPD v2 dataset. This implies that it can accurately replicate human judgement in judging synthetic pictures. The strength of HPSv2 is its strong training and ability to generalise across diverse types of pictures, making it the gold standard for evaluating image generating models.
\subsection{Inspirations to our project}
The methodologies discussed in these papers are directly related to the research question of how to optimize prompts for text-to-image tasks using LLMs without direct weight manipulation. Each method offers a unique perspective on managing the complexities of LLMs that are weight-inaccessible, providing various strategies from adversarial learning to reinforcement learning and textual gradients. These approaches are crucial for understanding how to effectively use LLMs as Prompt Modifiers, which can refine instructions given to another system, such as a text-to-image model, to enhance output quality.

The reviewed literature gives a detailed overview of several techniques for optimising prompts in text-to-image generation tasks. These strategies not only improve the direct outputs of such models, but they also widen the possible uses of LLMs in digital media development. By comparing diverse techniques, this review focuses on how academics are addressing the issues related with weight-inaccessible LLMs, opening the door for more advanced and accessible digital media tools.

\section{Methodology}
\subsection{prompt optimization for image generation}
In this project, we optimize the instruction of Generator, guiding it refine a naive prompt says \( X \) to the output \( Y \) which is an elaborate and refined prompt with detailed description in scene, color, style, quality,  succeeding in generating image correlating with original object and intention but more human preference fulfilling.

The architecture of the system is composed of three distinct language model agents operationalized via the GPT-3.5 Turbo, accessed through the OpenAI Assistant API. These agents are:

\begin{itemize}
  \item The \textbf{Generator} (\( \mathcal{G} \)), responsible for refine naive prompts, transforming them into detailed prompts suitable for generating high-fidelity images. The input to \( \mathcal{G} \) is represented by the tuple \( (I, X) \), where:
The output of \( \mathcal{G} \) is the refined prompt follows instruction.
\begin{itemize}
  \item \( I \) is the instruction undergoing optimization to enhance the capability of prompt generation.
  \item \( X \) is the naive prompt subject to refinement.
\end{itemize}

  \item The \textbf{Instruction Modifier} (\( \mathcal{I}_\text{mod} \)), which tasked with the modification and enhancement of the extant instructions. Its purpose is to incorporate specific improvements proposed by the Gradient Calculator (\( \mathcal{GC} \)) to elevate the quality and performance of generated prompts.

The operation of \( \mathcal{IM} \) is characterized by the input tuple \( (\text{Impr}, \text{Instr}) \), where:

\begin{itemize}
  \item \( \text{Impr} \) denotes the set of suggested improvements for the instruction, as determined by \( \mathcal{GC} \).
  \item \( \text{Instr} \) is the instruction that is currently yielding the least satisfactory results and is thereby targeted for refinement.
\end{itemize}

Upon receiving its inputs, \( \mathcal{IM} \) yields an output of \( n \) new instructions, each corresponding to \( n \) distinct improvements. Formally, this process is defined by the mapping \( \mathcal{IM}: (\text{Impr}, \text{Instr}) \mapsto \{\text{New\_Instr}_1, \text{New\_Instr}_2, \ldots, \text{New\_Instr}_n\} \), where \( \{\text{New\_Instr}_i\}_{i=1}^{n} \) represent the series of refined instructions.
  \item The \textbf{Gradient Calculator} (\( \mathcal{G}_\text{calc} \)), whose primary function is to deduce the incremental enhancements necessary for the Generator's instruction set. The rationale behind \( \mathcal{GC} \) is to analyze the differential in performance between prompts yielding lower scores and those with the highest scores within a given batch, which pertains to the same object—detailed further in the appendix.

The input for \( \mathcal{GC} \) is given by the triplet \( (\text{Instr}, \text{Low\_score\_prompt}, \text{High\_score\_prompt}) \), where:

\begin{itemize}
  \item \( \text{Instr} \) represents the current instruction correlated with the generation of the lowest average score prompts in the instruction list.
  \item \( \text{Low\_score\_prompt} \) is the set of prompts that, under the governance of \( \text{Instr} \), have resulted in the lowest average scores.
  \item \( \text{High\_score\_prompt} \) signifies the prompt evaluated as most human-preferred image in the prompts pool.
\end{itemize}

The output of \( \mathcal{GC} \) is a series of suggestions intended to refine the instruction, analogous to a gradient utilized in the iterative generation of new instructions.
\end{itemize}

Complementing these agents is HPS v2, a text-to-image evaluation model that gauges the affinity of generated images to human preferences. The model, symbolized by \( \mathcal{H} \), assigns scores to images crafted from diversified refined prompts, all stemming from the identical naive prompt \( X \). The mathematical representation of the scoring function is given by \( S: \mathcal{H} \times \mathcal{P} \rightarrow \mathbb{R} \), where \( \mathcal{P} \) represents the set of prompts, and \( \mathbb{R} \) denotes the real number scoring domain. Feedback from \( S \) informs subsequent refinements, ensuring adherence to a stringent criterion of aesthetic and pertinence in the generated images.

\begin{algorithm}
\caption{Algorithm of Generator prompt optimisation}
\begin{algorithmic}[1]

\State \textbf{Input:} \( G(\text{Instr}_i, x_i), GC(\text{Instr}_i, \text{low}_s \text{Prompt}_i, \text{High}_s \text{Prompt}_i), \)
\State \hspace{\algorithmicindent} \( IM(\text{Improvement}_i, \text{Instr}_i), S(\text{prompt}, \text{image}) \)
\State \textbf{where} \( G = \text{Generator}, GC = \text{Gradient Calculator}, \)
\State \hspace{\algorithmicindent} \( IM = \text{Instruction Modifier}, S = \text{Scorer} \)

\State \textbf{Initialization:} \( \text{instruction\_list} \gets [] \), \( \text{prompt\_pool} \gets \{\} \)

\For{\( \text{iter} \in \text{iterations} \)}
    \If{\( \text{instruction\_list} \text{ is empty} \)}
        \State \( \text{instruction\_list.append(init\_instr)} \)
    \EndIf
    \State \( \text{querys} \gets \text{sample}(\text{batch\_size naive prompts as querys}) \)
        \State \Comment{generate prompts based on existing instruction}
    \For{\( \text{instruction} \in \text{instruction\_list} \)}
        \State \( \text{batch\_loss} \gets 0 \)
        \For{\( \text{query} \in \text{querys} \)}
            \State \( \text{generated\_prompt} \gets G.\text{output} (\text{instruction}, \text{query}) \)
            \State \Comment{add generated prompt to prompt pool}
            \State \( \text{generated\_prompt } \to \text{prompts\_pool} \);
        
            \State \( \text{loss} \gets S(\text{query} \text{sdx\_turbo}(\text{generated\_prompt})) \)
            \State \( \text{batch\_loss} \gets \text{batch\_loss} + \text{loss} \)
        \EndFor
        \State \Comment{add instruction and its loss to list}
        \State \( \text{instruction\_list}[\text{instruction}] \gets (\text{Avg}(\text{batch\_loss}), \text{generated\_prompt\_list}) \)
    \EndFor
    
    \For{\( \text{query} \in \text{querys} \)}
    \State \Comment{add lexica\_prompts to prompt\_pool}
        \State \( \text{lexica\_prompts} \gets \text{fetch\_prompts\_from\_lexica}(\text{query}) \)
        
    \EndFor
    
    \State \( \text{epsilon\_greedy\_instr\_selection()} \)
    \State \( \text{instr\_to\_modify} \gets \text{find\_lowest\_instruction}() \)
    \State \( \text{improvement} \gets GC(\text{instr\_to\_modify}, X, \text{high\_score\_prompt}) \)
    \State \Comment{add new\_instr to instruction\_list}
    \State \( \text{new\_instr} \gets IM(\text{improvement}, \text{instr\_to\_modify}) \)
    \State \( \text{new\_instr} \to \text{instruction\_list} \);

\State \Comment{next iteration}
\EndFor
\end{algorithmic}
\end{algorithm}

\subsection{Batch Query Sampling}

To ensure our prompt Modifier generalizes across a wide range of user prompts, we adopt a strategy akin to batch gradient descent. In each iteration, we uniformly sample a batch of queries \( Q \) from the pool of simple prompts. The objective is to minimize the expected loss across all possible prompts, aiming for a Modifier that performs well on the average case rather than overfitting to specific instances.

Mathematically, we express this as optimizing the expected loss \( \mathbb{E}_{q \sim P} [L(f_{\theta}(q), y)] \), where \( \theta \) denotes the Modifier's parameters, \( L \) the loss function, \( f_{\theta} \) the Modifier function, \( q \) a sampled query, and \( y \) the target output.

By updating \( \theta \) based on the empirical loss from the batch, we regularize the training process to prevent overfitting, ensuring the Modifier's broad applicability.

\subsection{Selector}

In our methodology, the Selector component plays a critical role in maintaining a constant length of the instruction list, a challenge akin to the well-known Multi-Armed Bandit Problem. The reward metric in our context is the batch loss associated with each instruction in the tracks, providing a direct measure of instruction efficacy.

Our initial intention was to utilize the Upper Confidence Bound (UCB) algorithm, renowned for its balance between exploration of new strategies and exploitation of known rewarding actions. To validate this choice, we conducted comparative analyses with other prevalent selection strategies, including a greedy approach that consistently favors the instruction with the lowest batch loss, and an epsilon-greedy method, which introduces a probability of selecting a suboptimal instruction to allow for exploration beyond the immediately rewarding options.

\section{Results}
For witnessing the increase of quality of image generated using the optimized prompt of Generator, we set the baseline by using the normal instruction for prompt refinement and text\-image generation which is "This is the original prompt that you need to carefully refine, Prompt or subject to refine :\{query\}", the lexica prompt directed generated image and our optimised Generator instruction"Integrate exercises that challenge writers to distill their descriptions to the most essential elements while effectively evoking the desired emotions, reinforcing the lesson on brevity and precision in storytelling. Emphasize the use of impactful language and imagery to succinctly capture the essence of a scene and immerse the reader in a cohesive narrative that evokes awe, wonder, and exploration, ultimately igniting feelings of exhilaration and reverence for the boundless beauty and possibilities within the depicted setting,Prompt or subject to refine is : \{query\}"

\begin{figure}[htbp]
\centering
\begin{subfigure}[b]{0.3\textwidth}
  \includegraphics[width=\textwidth]{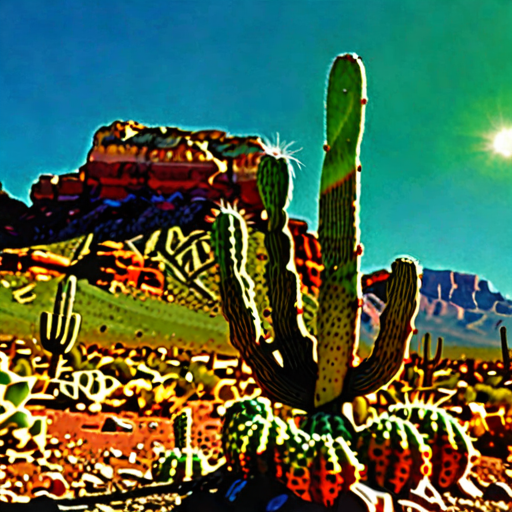}
  \caption{Baseline image generated using GPT-3.5's initial instruction}
  \label{fig:baseline}
\end{subfigure}
\hfill
\begin{subfigure}[b]{0.3\textwidth}
  \includegraphics[width=\textwidth]{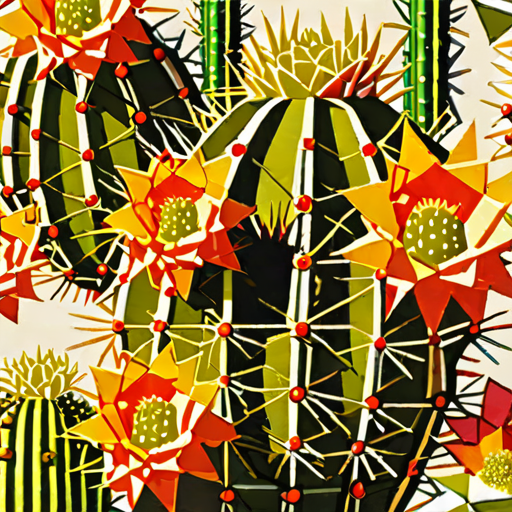}
  \caption{Image generated on Lexica's prompt}
  \label{fig:lexica}
\end{subfigure}
\hfill
\begin{subfigure}[b]{0.3\textwidth}
  \includegraphics[width=\textwidth]{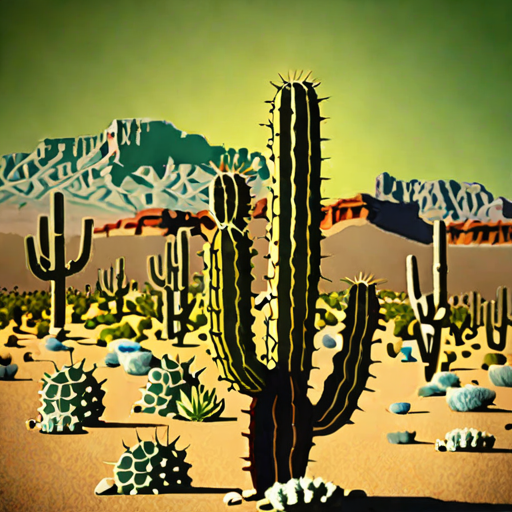}
  \caption{Optimized image using UCB selection method}
  \label{fig:ucb_optimized}
\end{subfigure}
\caption{Comparison of images generated from the same input "cactus" using different methods.}
\label{fig:comparison}
\end{figure}
The results of experiments are shown below in Table 1.

\begin{table}[H]
\label{results}
\centering
\caption{Results of experiments in different selection method and batch size.}
\begin{tabular}{l|c|c|c}
\toprule
\textbf{Batch size}&\textbf{Selection Method}& \textbf{initial Score}&\textbf{Last Iteration Score}  \\ \midrule
1&UCB&25.54& \textbf{28.78} \\
3&UCB&28.3& 27.69 \\
5&UCB&27.61&27.88 \\
3&greedy&27.62&28.02\\
3&epsilon\_greedy&27.98&28.72 \\
\bottomrule

\end{tabular}
\end{table}

\begin{table}[H]
\label{results}
\centering
\caption{Comparison with GPT3.5 and lexica Baseline}
\begin{tabular}{l|c|c|c}
\toprule
\label{Table 2}
\textbf{Method}&\textbf{Score}  \\ \midrule
GPT3.5 Baseline& 26.34 \\
Lexica Baseline& 26.14 \\
Our method& \textbf{26.72} \\
\bottomrule

\end{tabular}
\end{table}

able 1 reveals outcomes from testing different batch sizes and selection methods on image generation across 10 iterations. Using the Upper Confidence Bounds (UCB) method with a batch size of 1, scores improved from 25.54 to 28.78. However, a batch size of 3 with UCB saw a slight decrease from 28.3 to 27.69, while increasing to batch size 5 led to a small increase from 27.61 to 27.88. The greedy method with batch size 3 raised the score from 27.62 to 28.02, and the epsilon-greedy method increased it more significantly from 27.98 to 28.72.

These results indicate that smaller batch sizes with UCB showed the most significant improvement, whereas larger sizes under the same method slightly decreased or marginally improved. Conversely, both greedy and epsilon-greedy methods with a batch size of 3 enhanced scores, demonstrating how different strategies impact image generation efficacy.

Additionally, baseline comparisons \ref{Table 2} were conducted to evaluate our method against the GPT-3.5 baseline, where GPT-3.5 served as a prompt modifier, and Lexica, which acted as a human professional baseline by sourcing corresponding prompts based on queries. For fairness, 10 simple prompts were sampled, and average scores were extracted for each system. Results indicated that our method outperformed each baseline approach.

\subsection{Meaningful suggestion given by Gradient calculator}

The following are example outputs of the Gradient calculator, showcasing the low score and high score prompt group, followed by the inferred insights and proposed improvements for the instruction,guiding it to generate the structure and content approach high score prompts :

\begin{itemize}
  \item \textbf{Low Score Object 0:} Aquarium with sharks
  \begin{itemize}
    \item \textbf{Low Score Generated Prompt 0:} A captivating aquarium display featuring powerful sharks, offering a glimpse into the mysterious world of these majestic creatures in a controlled and mesmerizing environment.
    \item \textbf{Score:} 27.587890625
  \end{itemize}

  \item \textbf{Low Score Object 1:} Farm with windmill
  \begin{itemize}
    \item \textbf{Low Score Generated Prompt 1:} Enhanced Prompt: Immerse yourself in the idyllic charm of a countryside farm, where a majestic windmill towers over lush fields of swaying wheat and vibrant sunflowers...
    \item \textbf{Score:} 26.3671875
  \end{itemize}

  \item \textbf{High Score Object 0:} Aquarium with sharks
  \begin{itemize}
    \item \textbf{High Score Prompt 0:} An aquarium exhibit featuring majestic sharks gliding effortlessly through the water, their sleek bodies cutting through the vibrant underwater environment...
    \item \textbf{Score:} 28.5400390625
  \end{itemize}

  \item \textbf{High Score Object 1:} Farm with windmill
  \begin{itemize}
    \item \textbf{High Score Prompt 1:} Refined Prompt: A serene farm landscape featuring a classic windmill set against a backdrop of rolling fields, depicting the timeless charm of rural life and agricultural tradition.
    \item \textbf{Score:} 27.2705078125
  \end{itemize}
\end{itemize}

\subsection*{Inferences and Improvements}

\begin{itemize}
  \item \textbf{Inference 0:} The low score prompts lack depth and fail to evoke a sense of intrigue or engagement, missing opportunities to add layers of complexity or emotion to the scenes described.
  
  \item \textbf{Inference 1:} The low score prompts tend to be overly elaborate and excessively detailed, which can overwhelm the reader and detract from the core message of the prompt.
  
  \item \textbf{Improvement 0:} Emphasize the importance of creativity and originality in prompt generation, encouraging writers to think outside the box and incorporate unexpected elements to pique the reader's interest.
  
  \item \textbf{Improvement 1:} Encourage writers to strike a balance between detailed descriptions and concise communication in their prompts, ensuring that every detail adds value to the overall visualization without overwhelming the reader.
\end{itemize}
\section{Discussion}
The analysis presents a clear distinction in image quality when comparing outputs generated from initial and optimized instructions. The images crafted with our optimization methods demonstrably align more closely with human aesthetic preferences. This enhancement is reflected in our results table, where scores for images from the final iteration surpass those from the initial round, with the singular exception of the batch size three UCB experiment. The deviation in this instance is likely attributed to the exploratory component of the UCB algorithm, necessitating additional iterations to attain conclusive evidence of convergence and ultimate performance efficacy.

Adjustments to the batch size have revealed a correlation with the rate of convergence towards optimal instruction refinement, offering broader improvements across batches of prompts. However, this comes at the increased expense of input and output tokens for each iteration, indicating a trade-off between the rate of improvement and computational resource allocation.

Further, the study indicates that image quality is contingent upon the chosen diffusion model and the depicted object. For example, the SDX\_turbo model employed within our experiments demonstrates the capacity to generate visually appealing images with high scores, even from simpler, naive prompts. This observation is consistent across representations of specific subjects such as 'flaming Phoenix' and 'luxury yacht', where complex, detailed prompts do not necessarily yield a commensurate increase in image quality. These variations suggest that the perceived improvements in image scores from the last iteration of experiments may be less pronounced under certain conditions.
\section{Limitation}
The restrictions of this work are classified into three categories: time limits, computational resource limitations, and the unpredictability inherent in diffusion models.

The project's short plan prevented thorough experimentation with a variety of concepts. As a result, engaging with sophisticated models such as Llama or GPT-4 was not practical. Furthermore, model training was limited to 10 iterations at most, which is a significant constraint. Given the significant time and processing power necessary to fully train such models, the limited number of iterations most certainly hampered the models' capacity to achieve optimal convergence, as evidenced by the minor improvements seen in our studies.

Second, the exploratory scope of diffusion models was restricted. Diffusion models' performance can vary greatly—some models may create superior pictures with simple instructions, while others may require complex, comprehensive inputs to obtain equivalent results. As a result, a more in-depth examination into alternative diffusion models may have provided a more complete grasp of their possibilities.

For future goals, if greater computer resources become available, the number of training iterations should be significantly increased. Furthermore, doing a comparison analysis across a range of diffusion models is likely to broaden and deepen the findings.

\section{Code}
The github link is given here, please click on it to reach our repository : \href{https://github.com/slark-prime/MAAI_Optimizer}{Group20-MAAI-Code}
\appendix
\newpage
\section{Example Instructions}

{\small
\begin{longtable}{l|p{12cm}}\hline
    \hline \multicolumn{2}{c}{\textbf{Prompt Design}} \\ \hline 
    \endfirsthead
    
    \multicolumn{2}{c}%
    {{\bfseries \tablename\ \thetable{} -- continued from previous page}} \\
    \hline \multicolumn{2}{c|}{\textbf{Prompt Design}} \\ \hline 
    \endhead
    
    \hline \multicolumn{2}{r}{{Continued on next page}} \\ \hline
    \endfoot

    \endlastfoot

    \multicolumn{2}{l}{\textbf{Demonstration}} \\* \hline
    Gradient Calculator Input & Analyze the following low score and high score batch, each prompt with corresponding scores. And infer what's wrong with the instruction generating low score batch prompt to suggest the improvement of the instruction:For your answer use the format:
    
    Inference 1: your inference\_1
    
    Inference 2: your inference\_2 
    
    Inference n: your inference\_n... 
    
    Improvement 1: you suggested improvement correspond to inference 1
    
    Improvement 2: you suggested improvement correspond to Inference 2
    
    Improvement n: you suggested improvement correspond to inference n...
    
    This is the generator instruction:\{instruction\} and first corresponding generated low score prompts group: 
    
    low\_score\_object\{i\}:\{obj\_i\}, 
    
    low\_score\_generated\_prompt:
    
    \{low\_score\_prompt\_i\},score:\{low\_score\_i\}

    below is high score prompts group:
    
    high\_score\_object\{i\}:\{obj\_i\}, 
    
    high\_score\_prompt:
    
    \{high\_score\_prompt\_i\},score:
    
    \{high\_score\_i\}"\\* \hline

    Gradient Calculator Output & analyze\_and\_propose: Summary of Reasons for Scores:
    
    Inference 1: The low score prompts may have focused too much on describing the hot air balloon and the view, while not incorporating the feeling of exhilaration and awe effectively.
    
    Inference 2: The low score prompts may have lacked creativity in conveying the emotions of exhilaration and awe through the scene described.
    
    Improvement 1: Encourage writers to not only describe the setting but also focus on capturing the emotions of exhilaration and awe experienced by the passengers in the hot air balloon.
    
    Improvement 2: Provide examples or suggestions on how to infuse more creativity and emotional depth into the description of the panoramic view, making it more evocative of exhilaration and awe.
    
    Improvement 3: Include guidance on using vibrant language and sensory details to intensify the reader's experience of exhilaration and awe in the scene described.\\* \hline
\bottomrule
    
\caption{Examples of Gradient Calculator Input and Output}
    \label{table:state-prompt}
\end{longtable}
}

\bibliographystyle{plainnat}
\bibliography{reference}

\end{document}